\documentclass[11pt]{article}

\usepackage[margin=1in]{geometry}
\usepackage{amsmath,amssymb}
\usepackage{graphicx}
\usepackage{array}
\usepackage{algorithm}
\usepackage{algpseudocode}
\usepackage{changepage}      
\usepackage{rotating}        
\usepackage{afterpage}
\usepackage[aboveskip=1pt,labelfont=bf,labelsep=period,justification=raggedright,singlelinecheck=off]{caption}
\usepackage{cite}
\usepackage{hyperref}

\newlength\savedwidth
\newcommand\thickhline{\noalign{\global\savedwidth\arrayrulewidth\global\arrayrulewidth 2pt}%
\hline
\noalign{\global\arrayrulewidth\savedwidth}}

\title{\bf EMFE: A lightweight, explainable machine learning framework for malaria cell classification}

\author{
Md Abdullah Al Kafi$^{1}$,
Walayat Hussain$^{3}$,
Mousumi Karmakar$^{2}$,
\\
Sumit Kumar Banshal$^{2}$,
Ahmed Al Marouf$^{4,\ast}$
\\[3mm]
\small $^{1}$Multidisciplinary Action Research (MARS) Lab, Daffodil International University, Dhaka, Bangladesh\\
\small $^{2}$Department of Computer Science and Engineering, Alliance University, Bangalore, India\\
\small $^{3}$Artificial Intelligence for Decision Excellence (AIDX) Lab, Peter Faber Business School,\\
\small \phantom{$^{3}$}Australian Catholic University, North Sydney, Australia\\
\small $^{4}$Department of Medicine, Faculty of Medicine \& Dentistry, University of Alberta, Edmonton, Canada\\[2mm]
\small $^{\ast}$Corresponding author: \texttt{ahmed.marouf@ualberta.ca}
}
\date{}

\begin{document}
\maketitle

\begin{abstract}
Automated malaria diagnosis from stained blood-smear microscopy is dominated by deep convolutional neural networks that are accurate but computationally expensive, poorly interpretable, and rarely validated with patient-level rigor. We present EMFE (Efficient Mathematical Feature Extraction), a five-feature mathematical feature-extraction framework for classifying single red-blood-cell images as parasitized or uninfected, combining Gray World color normalization, adaptive green-channel thresholding, and morphological spot detection with classical machine learning classifiers. Using the National Institutes of Health (NIH) Lister Hill National Center for Biomedical Communications (LHNCBC) malaria dataset (27{,}558 images, 200 patients), we evaluate three classifiers, namely Random Forest, Histogram Gradient Boosting, and Support Vector Machine, under a patient-grouped nested cross-validation protocol ($K_\text{outer}=20$, $K_\text{inner}=3$) that keeps every patient's cells confined to a single fold, eliminating a leakage risk common in this literature. The optimized Random Forest achieves 94.6\% pooled out-of-fold accuracy (95\% CI [93.6, 95.7]), corroborated by an untouched 40-patient holdout test (94.3\%) and a patient-level permutation test ($p<0.001$, 1{,}000 permutations). A controlled ablation isolates each feature's and each pipeline stage's contribution. A hardware-matched comparison against retrained DenseNet121, ResNet50, and MobileNetV2 baselines quantifies the accuracy-efficiency trade-off with a properly powered paired statistical test. A synthetic imaging-perturbation study characterizes three specific failure modes, and a full explainability analysis identifies spot saturation as the dominant discriminative feature. We further report a patient-level aggregation analysis quantifying the sensitivity-specificity trade-off and false-positive accumulation when per-cell predictions are pooled into a per-patient call. Together, these results provide a statistically rigorous, mathematically transparent, and computationally lightweight alternative to deep learning for this task, with quantified limitations on where that evidence holds.
\end{abstract}

\section{Introduction}

Malaria remains one of the most burdensome mosquito-borne diseases worldwide, concentrated in Sub-Saharan Africa and Southeast Asia, where the causative \textit{Plasmodium} parasite continues to cause hundreds of thousands of deaths annually~\cite{Mairi2025}. The reference diagnostic standard, microscopic examination of Giemsa-stained thin blood smears, is accurate but slow, requires a trained microscopist, and is difficult to scale in the low-resource settings where malaria transmission is highest~\cite{Poostchi2018}. These constraints have motivated a large body of work on automated, image-based malaria cell classification.

The dominant approach in this literature is deep convolutional neural networks (CNNs)~\cite{Krizhevsky2017}, which learn discriminative features directly from raw pixel data and, on the NIH LHNCBC single-cell dataset used throughout this study, routinely report accuracies in the high 90s~\cite{Rajaraman2018}. This accuracy, however, comes with three recurring costs that are the specific motivation for the present work.

First, CNNs are computationally expensive, requiring large model files, substantial memory, and, typically, GPU acceleration for practical inference throughput. Diagnostic settings in malaria-endemic regions frequently lack reliable power, GPU hardware, or stable connectivity, making GPU-dependent pipelines difficult to deploy at the point of care~\cite{Yang2019}.

Second, CNNs are largely uninterpretable. A prediction is a function of millions of learned weights with no direct correspondence to a clinically meaningful quantity, which limits the trust a health worker can place in an individual prediction and complicates auditing a model's failure modes~\cite{Lipton2018}.

Third, and most consequentially for the validity of reported accuracy figures, the NIH dataset, and datasets like it, contain multiple cell images per patient. A cross-validation split performed at the image level, without grouping by patient, allows a given patient's cells to appear in both the training and test partitions of the same fold. A model can then partially learn patient- or slide-specific characteristics (staining batch, illumination, imaging artifacts) rather than transferable morphological features, inflating the apparent accuracy relative to what a model would achieve on an unseen patient. Two of the very few studies in this literature that explicitly measure this effect directly report the difference: cell-level accuracy of 98.6\% falling to 95.9\% at the patient level in one case~\cite{Rajaraman2018}, and 96.9\% falling to 78.0\% in another~\cite{Yu2020}. Many published classical-feature malaria-classification studies do not report whether their evaluation was patient-grouped, which makes it difficult to know from the published record alone how much of their headline accuracy reflects generalization versus partial leakage.

Classical, handcrafted-feature approaches offer a potential answer to the first two costs: hand-engineered morphological, textural, and color features are inexpensive to compute, run in milliseconds on a CPU, and, because each feature has an explicit mathematical definition, are directly inspectable~\cite{Ross2006,Das2013,Devi2016}. This literature has generally lacked rigor on the third point: patient-grouped evaluation, adequately powered statistical significance testing, and a controlled, hardware-matched comparison against deep-learning alternatives rather than an assumed or partial one. A common substitute for proper significance testing in this literature is a one-sample test against a chance baseline, which incorrectly treats non-independent cross-validation folds as independent observations.

We address these three gaps with a five-feature mathematical descriptor set, Efficient Mathematical Feature Extraction (EMFE), built around Gray World color normalization and adaptive green-channel spot detection. It substantially extends our own preliminary, non-peer-reviewed investigation of this general idea~\cite{Kafi2025Preprint}, which used a minimal two-feature descriptor evaluated with an image-level, non-patient-grouped train/test split and a single significance test. Here we expand to a five-feature descriptor set with an explicit biological rationale for each feature, replace the image-level split with the patient-grouped nested cross-validation protocol described below, and add the full statistical validation stack, ablation study, robustness characterization, explainability analysis, and patient-level aggregation analysis that the preliminary work lacked. Beyond fixing the specific statistical and leakage problems identified above, we provide the full mathematical formulation and biological rationale for every processing stage and every feature, a companion ablation study isolating each feature's and stage's individual contribution, a synthetic robustness characterization, a full explainability analysis, and a patient-level aggregation analysis. Collectively, these form a methodologically complete account of what this feature set does, why each design choice was made, and precisely where its validated evidence ends.

\subsection*{Contributions}

This study makes the following contributions:
\begin{itemize}
    \item A five-feature mathematical descriptor set for single-cell malaria classification, each feature given an explicit biological or optical rationale grounded in Giemsa-stain chemistry rather than the millions of opaque parameters a CNN would use for the same task.
    \item A patient-grouped, nested cross-validation protocol ($K_\text{outer}=20$, $K_\text{inner}=3$) with a runtime leakage guard, showing that closing the leakage gap this literature's own prior work has measured at 3--19 accuracy points~\cite{Rajaraman2018,Yu2020} still yields 94.6\% pooled accuracy, corroborated on an untouched 40-patient holdout.
    \item A statistical validation stack confirming that accuracy is not attributable to chance (patient-level permutation test, $p<0.001$) and holds under patient-cluster-bootstrap resampling on an untouched holdout set.
    \item A controlled ablation isolating spot saturation as the single dominant discriminative feature and showing the full five-feature design outperforms four classical baseline descriptors by 1.5--29 accuracy points.
    \item A hardware-matched comparison showing EMFE trails fully retrained DenseNet121, ResNet50, and MobileNetV2 by only 1.7--2.4 accuracy points while running 3.8--43$\times$ faster and 1.4--14.3$\times$ smaller on identical CPU hardware, the first matched quantification of this trade-off against both large and lightweight architectures in this literature.
    \item A synthetic robustness characterization identifying three specific perturbations that collapse specificity while sensitivity stays high, rather than causing diffuse, unexplained failure.
    \item An explainability analysis in which three independent methods, ablation, permutation importance, and failure-mode comparison, converge on spot saturation as the dominant signal, evidence that this reflects a genuine data property rather than an analysis artifact.
    \item A patient-level aggregation analysis showing naive per-cell aggregation is unusable in practice (14\% control-cohort specificity), while a modest threshold recovers 100\% specificity at a 12.7-point sensitivity cost concentrated in low-parasitemia patients.
\end{itemize}

The remainder of this paper is organized as follows. Section~\ref{sec:methods} describes the dataset, the full mathematical feature-extraction pipeline, the classifiers and their optimization, and the design of every validation, ablation, robustness, comparison, explainability, and aggregation analysis. Section~\ref{sec:results} reports the outcome of each. Section~\ref{sec:discussion} interprets these findings jointly and states the study's limitations explicitly, followed by a brief conclusion.

\section{Background}

\subsection{Deep Learning Approaches to Malaria Cell Classification}

Over the past decade, deep learning, primarily driven by Convolutional Neural Networks (CNNs), has emerged as the dominant methodology for automated microscopic malaria cell image classification \cite{Poostchi2018,Rajaraman2018,Yang2019}. Researchers have evaluated a broad spectrum of architectures ranging from custom-designed deep networks to large pre-trained models fine-tuned via transfer learning \cite{Rajaraman2018,Krizhevsky2017}. For instance, Liang et al. proposed a custom 16-layer CNN architecture trained on single-cell images from the public National Institutes of Health (NIH) Lister Hill National Center for Biomedical Communications (LHNCBC) repository, reporting a classification accuracy of 97.37\% under 10-fold cross-validation and demonstrating clear superiority over un-tuned transfer learning baselines \cite{Liang2016}. Similarly, Bibin et al. formulated a 6-layer Deep Belief Network (DBN) for peripheral blood smear images, achieving an accuracy of 96.40\% on a dataset of 4,100 single cells \cite{Bibin2017}.

Other studies have leveraged transfer learning from large-scale natural image databases. Rajaraman et al. evaluated six pre-trained CNN architectures, including AlexNet, VGG-16, and ResNet-50, demonstrating that extracting features from optimal intermediate layers yielded cell-level classification accuracies reaching up to 98.61\% \cite{Rajaraman2018}. Further empirical evaluations by Loddo et al. confirmed that modern deep architectures like DenseNet and ResNet consistently achieve high classification performance exceeding 97\% on benchmark single-cell datasets \cite{Loddo2022}.

However, these headline accuracy figures in the high 90s come at a substantial cost \cite{Yang2019}. Deep CNNs require significant computational infrastructure, rely heavily on dedicated Graphics Processing Units (GPUs) for timely inference, and operate as opaque black boxes whose millions of parameters lack direct mathematical or biological mapping to clinical visual cues \cite{Lipton2018}. Crucially, as discussed in subsequent subsections, most of these deep learning studies rely on randomized image-level evaluations that leave open the risk of unexamined patient-level data leakage \cite{Rajaraman2018,Yu2020}.

\subsection{Classical and Handcrafted-Feature Approaches}

Prior to the dominance of deep neural networks, computational malaria diagnostics relied on classical computer vision pipelines combining handcrafted mathematical descriptors with traditional machine learning classifiers, such as Support Vector Machines (SVM), Random Forests (RF), and Naive Bayes \cite{Ross2006,Das2013,Devi2016}. These pipelines systematically extract explicit visual properties across three main feature domains: morphology (such as cell area, perimeter, and circularity), color distributions (such as transformations from RGB to HSV, YCbCr, or $L^*a^*b^*$ color spaces), and local texture metrics \cite{Das2013,Devi2016}. Textural representations frequently incorporate Gray-Level Co-occurrence Matrices (GLCM), Local Binary Patterns (LBP), or Histograms of Oriented Gradients (HOG) to capture local intensity variations \cite{Haralick1973,Ojala2002,Dalal2005}.

Classical handcrafted approaches possess distinct operational advantages: they are computationally cheap to evaluate, run efficiently on Central Processing Units (CPUs) without requiring specialized hardware, and offer complete mathematical transparency \cite{Ross2006}. Nevertheless, this literature has generally not matched deep learning studies in validation rigor \cite{Poostchi2018}. Studies in this paradigm have typically not reported patient-grouped cross-validation, adequately powered non-parametric statistical significance testing, or controlled component-wise ablation analyses to isolate the biological sources of predictive performance \cite{Poostchi2018}.

\subsection{The Patient-Level Leakage Problem}

A common limitation in much of the automated malaria microscopy literature is the omission of patient-grouped cross-validation \cite{Rajaraman2018,Yu2020}. Public benchmark repositories, most notably the NIH LHNCBC dataset containing 27,558 single-cell images, are constructed by extracting multiple cell patches from blood smear slides originating from a limited pool of 200 distinct patients \cite{Rajaraman2018}. Cells extracted from the same patient slide share identical visual confounders, including specific Giemsa staining concentrations, local slide illumination gradients, artifact density, and camera sensor noise \cite{Poostchi2018}. When cross-validation splits are performed randomly at the single-cell image level rather than grouped by patient identity, single cells from the same patient slide appear simultaneously in both the training and test partitions \cite{Rajaraman2018}. Under these un-grouped conditions, a classifier can easily memorize slide-specific background signatures rather than learning genuine intra-erythrocytic parasite morphology, resulting in artificially inflated cross-validation performance \cite{Yu2020}.

That data leakage substantially alters reported performance is not a theoretical speculation; it is directly quantified by the very few studies in this literature that explicitly compared cell-level versus patient-level evaluation protocols \cite{Rajaraman2018,Yu2020}. Rajaraman et al. explicitly documented this drop, observing model classification accuracy fall from a cell-level figure of 98.61\% down to 95.90\% when evaluated at the patient level \cite{Rajaraman2018}. Yu et al. reported a larger drop, from 96.90\% accuracy under cell-level evaluation to 78.00\% under patient-grouped evaluation \cite{Yu2020}. These documented drops indicate that un-grouped evaluations can substantially overstate model generalization to unseen clinical patients \cite{Yu2020}.

\subsection{Statistical Rigor and Validation Practices in This Literature}

Beyond evaluation splitting errors, published literature in automated malaria diagnosis frequently exhibits weaknesses in statistical validation and reporting practices \cite{Poostchi2018}. A common deficiency is the reliance on isolated point estimates without reporting confidence intervals, or the inappropriate application of standard parametric tests, such as one-sample $t$-tests against chance or arbitrary baselines, that treat non-independent cross-validation folds as independent observations \cite{Demsar2006}. Furthermore, many studies report performance exclusively on cross-validation folds without reserving an untouched, patient-disjoint holdout test set to corroborate generalization post-optimization \cite{Rajaraman2018}.

Establishing diagnostic reliability with confidence requires a more complete statistical validation stack \cite{Demsar2006}. This includes employing patient-level non-parametric permutation testing to verify performance against global label shuffling while preserving patient-grouped fold structure, deriving patient-cluster-bootstrap confidence intervals on unseen holdout test sets, and utilizing paired non-parametric statistical tests, such as the paired Wilcoxon signed-rank test, to compare classifiers across identical cross-validation folds \cite{Demsar2006}.

\subsection{Interpretability and Computational Cost Trade-Offs}

The real-world deployment of automated diagnostic tools in endemic regions, such as Sub-Saharan Africa and Southeast Asia, is shaped by substantial practical constraints \cite{Yang2019,Mairi2025}. Point-of-care clinics in low-resource settings frequently lack reliable electrical power, high-performance computing infrastructure, and dedicated GPU hardware \cite{Yang2019}. Modern deep learning models, while highly accurate, feature large memory footprints occupying tens to hundreds of megabytes and high inference latencies on CPU hardware, which limits edge deployment on portable devices or mobile microscopes \cite{Yang2019}.

Some prior work has targeted this constraint directly rather than relying on large general-purpose architectures. Eze and Asogwa quantify the accuracy/size trade-off of INT8-quantized CNNs for malaria classification, reducing model size to 2~MB with microsecond-scale mobile inference latency at 97--98\% accuracy~\cite{Eze2021}. Nakasi et al.\ target mobile-deployable parasite and white-blood-cell localization on thick smears using quantized object-detection models~\cite{Nakasi2021}. Neither study reports patient-grouped evaluation, leaving both exposed to the same leakage risk described above for the broader literature.

In addition to computational overhead, the opaque black-box nature of deep neural networks is a further obstacle to clinical adoption \cite{Lipton2018}. Because a CNN's prediction is derived from non-linear combinations across millions of learned parameters, health workers cannot easily verify whether a decision is based on true parasite chromatin structures or peripheral staining artifacts \cite{Lipton2018}. This opacity limits clinical trust and complicates auditing model failure modes \cite{Lipton2018}. This motivates lightweight, mathematically transparent feature-extraction frameworks that execute rapidly on standard CPU hardware while maintaining rigorous, statistically validated diagnostic accuracy \cite{Ross2006,Yang2019}.

\subsection{Summary of Gaps and Positioning of the Present Study}

A synthesis of the literature reveals three major unaddressed gaps across existing studies:
\begin{enumerate}
    \item \textbf{Data Leakage and Validation Deficits}: Deep learning models achieve headline accuracies in the high 90s, but often use un-grouped cell splits that can induce patient-level data leakage, without rigorous non-parametric statistical testing or untouched patient holdout validation \cite{Rajaraman2018,Yu2020}.
    \item \textbf{Interpretability and Deployment Overhead}: Classical handcrafted approaches offer low computational cost and inspectability, but historically lack patient-grouped validation, structured component ablations, and matched deep learning benchmarks \cite{Ross2006,Poostchi2018}.
    \item \textbf{Unverified Trade-Offs}: No single prior study provides a hardware-matched, statistically validated comparison that directly quantifies the exact accuracy-for-efficiency trade-off between lightweight mathematical features and deep learning architectures evaluated under identical, leakage-free patient splits \cite{Poostchi2018}.
\end{enumerate}

The Efficient Mathematical Feature Extraction (EMFE) framework introduced here combines a deterministic 5-feature descriptor set with a patient-grouped nested cross-validation protocol, a complete statistical validation stack built on established non-parametric methodology~\cite{Demsar2006}, rigorous ablation and failure-mode analyses, and hardware-matched comparisons against retrained deep learning baselines. Table~\ref{tab:literature_comparison} provides a systematic comparative summary of landmark studies alongside the proposed EMFE framework across these core design dimensions.

\afterpage{%
\begin{sidewaystable*}[p]
\centering
\caption{\textbf{Systematic methodological comparison of automated malaria cell classification literature alongside the EMFE framework.}}
\label{tab:literature_comparison}
\resizebox{\textheight}{!}{%
\small
\begin{tabular}{|p{1.3in}|p{1.1in}|c|p{1.2in}|p{0.9in}|c|p{1.0in}|p{1.0in}|}
\hline
\textbf{Study} & \textbf{Feature type} & \textbf{Patient-grouped CV} & \textbf{Cell$\rightarrow$patient accuracy reported} & \textbf{Stat. test} & \textbf{Ablation} & \textbf{DL baseline (matched)} & \textbf{Explainability} \\
\hline
\textbf{Liang et al. (2016/2017)} \cite{Liang2016} & Deep CNN (16-layer custom) & No (10-fold image split) & Cell: 97.37\% / Patient: Not reported & None (point estimates) & No & Yes (AlexNet transfer: 91.99\%) & None (black box) \\
\hline
\textbf{Bibin et al. (2017)} \cite{Bibin2017} & Deep Belief Net (DBN) & No (Random cell split) & Cell: 96.40\% / Patient: Not reported & None & No & None & None \\
\hline
\textbf{Rajaraman et al. (2018)} \cite{Rajaraman2018} & Pre-trained CNNs (VGG/ResNet) & Partial (discussed split) & Cell: 98.61\% / Patient: 95.90\% & ANOVA / Kruskal-Wallis & No & Yes (6 pre-trained CNNs) & Activation maps \\
\hline
\textbf{Yu et al. (2020)} \cite{Yu2020} & Deep CNN (Thin/Thick) & Yes & Cell: 96.90\% / Patient: 78.00\% & Two-sample $t$-test & No & No & None \\
\hline
\textbf{Loddo et al. (2022)} \cite{Loddo2022} & Deep CNNs (VGG/ResNet) & No (Random cell split) & Cell: $\sim$97.00\% / Patient: Not reported & None & No & Yes (Multi-CNN eval) & Grad-CAM \\
\hline
\textbf{EMFE Framework (Present Study)} & Handcrafted 5-feature vector & Yes ($K_{\text{outer}}=20$, $K_{\text{inner}}=3$ nested) & Cell: 94.63\% (Pooled OOF), 94.29\% (Holdout) / Patient: Evaluated ($\tau=1\dots20$) & Permutation ($p<0.001$), Wilcoxon ($p<10^{-5}$) & Yes (LOFO + 4 stage + 4 baselines) & Yes (DenseNet121, ResNet50, MobileNetV2) & Permutation importance, PDP, Mann-Whitney $U$ \\
\hline
\end{tabular}%
}
\end{sidewaystable*}
\clearpage
}

\section{Materials and methods}
\label{sec:methods}

\subsection{Dataset and patient structure}

This study uses the publicly available NIH LHNCBC malaria single-cell dataset~\cite{MalariaProject,Rajaraman2018}, which contains 27{,}558 microscopic images of individual red blood cells extracted from Giemsa-stained thin blood-smear slides, labeled \textit{Parasitized} or \textit{Uninfected}. The dataset is exactly class-balanced (13{,}779 images per class). NIH additionally provides official patient-to-image mapping files, which we use as the ground truth for patient identity throughout this study: 150 \textit{P.\ falciparum}-infected patients, whose slides contribute cells to both classes (an infected patient's slide is a mixture of parasitized and uninfected-looking cells), and 50 additional uninfected-only control patients, for 200 distinct patients in total. Every image's filename-embedded patient token was independently cross-checked against the official mapping and agreed in all 27{,}558 cases. All 27{,}558 images were successfully decoded and processed; none were excluded.

Every cross-validation split performed in this study, inner and outer, across every analysis described below, is grouped by this patient identity, so that no patient's cells can appear in both the training and test partition of the same fold. This is the direct methodological response to the leakage risk described in the Introduction, and is enforced not only by construction of the splits but by an explicit runtime assertion (Section~\ref{sec:nestedcv}) that halts execution if any patient ID is found in both partitions of any fold.

\subsection{Mathematical feature-extraction pipeline}
\label{sec:pipeline}

\begin{figure}[!h]
\centering
\includegraphics[width=0.65\linewidth]{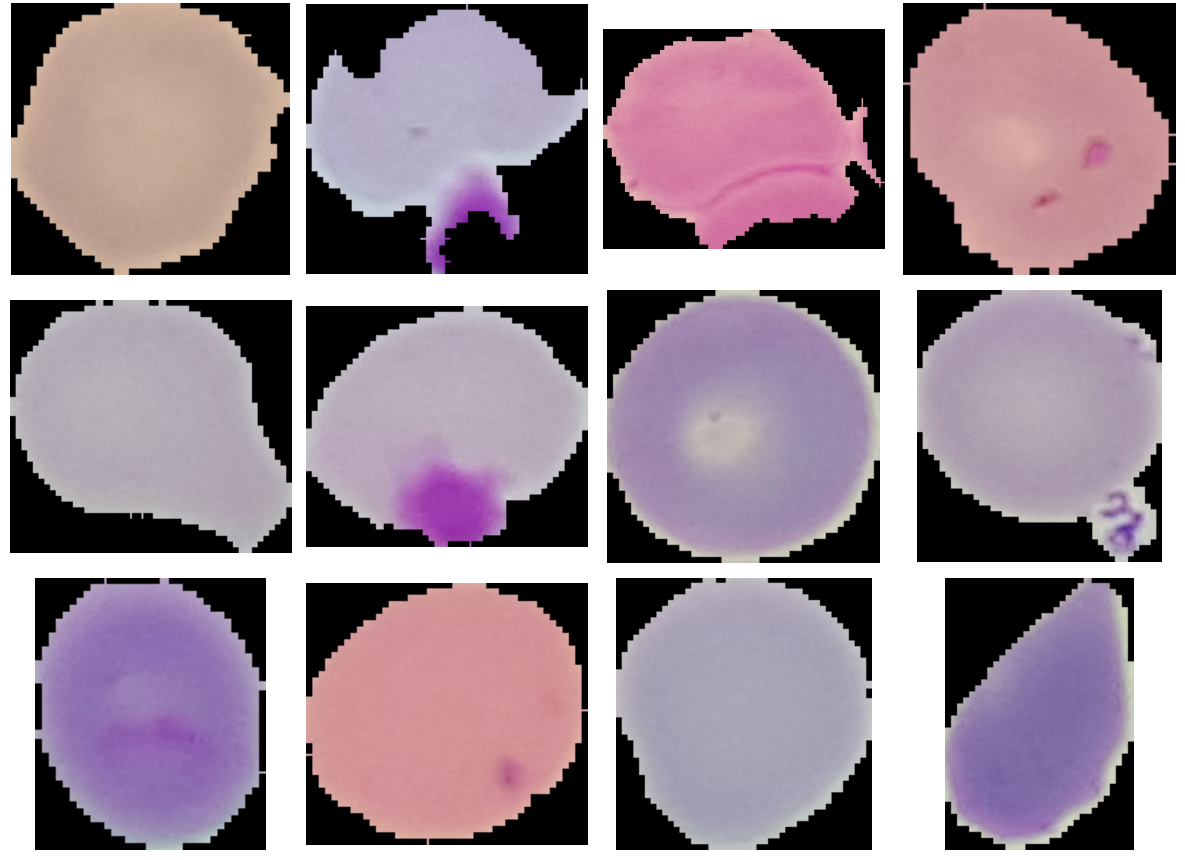}
\caption{\textbf{Representative sample images from the dataset.} Parasitized cells (containing visible \textit{Plasmodium} chromatin, ring forms, or trophozoites) and uninfected erythrocytes, illustrating the range of staining intensity, illumination, and background artifact present in the raw data.}
\label{fig:dataset}
\end{figure}

Raw cell images vary substantially in stain concentration, illumination, resolution, and aspect ratio (Fig~\ref{fig:dataset}). Before feature extraction, every image is resized to $128\times128$ pixels. The pipeline that follows is fully deterministic: every quantity below is a closed-form function of the input image, with no learned parameters, so a given image always produces the same feature vector and every step of the computation can be inspected directly.

\paragraph{Color normalization.} Stain concentration and illumination differ substantially between slides and imaging sessions, shifting the absolute color of otherwise similar cells. We correct for this with Gray World color normalization, which rescales each color channel so that the image's overall average color becomes neutral. Given an image with per-channel means $\mu_B, \mu_G, \mu_R$ over the blue, green, and red channels and their average $\mu = (\mu_B + \mu_G + \mu_R)/3$, each channel $C \in \{B, G, R\}$ is rescaled as
\begin{equation}
\label{eq:graycorld}
C' = \mathrm{clip}\!\left(C \cdot \frac{\mu}{\mu_C},\ 0,\ 255\right).
\end{equation}
This is a standard illuminant-correction assumption: under the Gray World hypothesis, deviation of a channel's mean from the overall gray level is attributed to a color cast (staining or illumination) rather than genuine image content, and Eq~(\ref{eq:graycorld}) removes that cast while preserving relative contrast within the image. If any channel mean is exactly zero, a degenerate case corresponding to an image that is fully black in that channel, normalization is skipped for that image and the original image is used unmodified; the sample is retained, not silently dropped, and the occurrence is logged (this never occurred in the 27{,}558-image dataset used here, but the pipeline handles it explicitly rather than dividing by zero).

\paragraph{Cell mask.} The color-normalized image is converted to grayscale ($I_\text{gray}$), and a binary mask separating the cell from the black background is obtained by a fixed global threshold,
\begin{equation}
M_\text{cell} = \mathbb{1}[I_\text{gray} > 10],
\end{equation}
exploiting the fact that the background in this dataset is uniformly near-black while any cell tissue is substantially brighter.

\paragraph{Spot detection.} Candidate parasite signal is sought in the green channel of the normalized image ($G'$), which provides higher contrast for stained chromatin than the red or blue channels under Giemsa staining. Rather than a single global threshold, we apply adaptive Gaussian thresholding, which computes a locally-varying threshold from each pixel's neighborhood:
\begin{equation}
T(x, y) = \mathbb{1}\!\left[G'(x, y) < \mathcal{N}_{b}(x,y) - C\right],
\end{equation}
where $\mathcal{N}_b(x,y)$ is the Gaussian-weighted mean of $G'$ over a $b \times b$ neighborhood centered at $(x,y)$ (default $b=21$), and $C$ (default $5$) is a constant subtracted from that local mean before comparison. Local adaptive thresholding is used, rather than a single global cutoff, because parasite-bearing regions are locally darker than their immediate surroundings even when overall cell brightness varies across the image (e.g. due to residual staining gradients uncorrected by Eq~\ref{eq:graycorld}); a global threshold would either miss faint local spots on bright cells or over-detect on dark cells. The raw threshold response is restricted to the cell interior, $S_\text{raw} = T \cap M_\text{cell}$, and cleaned with morphological opening (erosion followed by dilation) using a $3\times3$ elliptical structuring element $K$,
\begin{equation}
S_\text{clean} = (S_\text{raw} \ominus K) \oplus K,
\end{equation}
which removes single-pixel noise from the thresholded mask without eroding genuine, spatially coherent spot regions.

\paragraph{The five features.} Connected components of $S_\text{clean}$ are extracted as contours $\{c_1, \dots, c_{N}\}$. The final feature vector $\mathbf{F} \in \mathbb{R}^5$ is:
\begin{itemize}
    \item $N_\text{spots}$: the number of detected connected components, a coarse proxy for how many spatially distinct candidate dark regions are present in the cell.
    \item $A_\text{max} = \max_i \text{area}(c_i)$: the pixel area of the largest connected component. Larger, more developed parasite structures (e.g.\ trophozoites) are expected to occupy a larger contiguous area than staining artifacts or noise.
    \item $A_\text{total} = \sum_i \text{area}(c_i)$: the summed pixel area of all components, a proxy for total candidate parasite burden within the cell that is robust to whether that signal is concentrated in one region or fragmented across several.
    \item $S_\text{max}$: the mean HSV saturation of the pixels within the largest component, $S_\text{max} = \text{mean}\!\left(I_\text{HSV}^{(S)}\big[\,M_{c_{\max}}=1\,\right])$, where $c_{\max} = \arg\max_i \text{area}(c_i)$ and $I_\text{HSV}^{(S)}$ is the saturation channel of the (normalized) image in HSV space. Giemsa-stained chromatin and hemozoin pigment absorb stain intensely and appear as strongly saturated purple/magenta relative to the comparatively pale, desaturated cytoplasm of an uninfected erythrocyte; saturation is accordingly the feature we expect, and later confirm empirically (Section~\ref{sec:res-ablation}), to carry the strongest discriminative signal.
    \item $\sigma_\text{gray}$: the standard deviation of grayscale intensity within the cell mask, $\sigma_\text{gray} = \text{std}(I_\text{gray}[M_\text{cell}=1])$, a coarse measure of overall textural heterogeneity within the cell that captures variation not localized to a single connected spot.
\end{itemize}
If no connected component is found ($N=0$), $A_\text{max} = A_\text{total} = S_\text{max} = 0$ by definition and $\sigma_\text{gray}$ is computed as above over the whole cell mask. All five quantities are raw, non-normalized pixel-area or intensity values computed over the $128\times128$ resized image, not rescaled by cell size; this design choice is examined directly in the ablation study.

Algorithm~\ref{alg:feature_extraction} summarizes the complete pipeline.

\begin{algorithm}
\caption{Cellular feature extraction pipeline}
\label{alg:feature_extraction}
\begin{algorithmic}[1]
\Require Image path $I_\text{path}$, target size $S=128\times128$
\Ensure Feature vector $\mathbf{F}\in\mathbb{R}^5$
\State $I \gets \text{LoadImage}(I_\text{path})$; \textbf{if} $I$ is null \textbf{return} null
\State $I \gets \text{Resize}(I, S)$
\State $B,G,R \gets \text{SplitChannels}(I)$
\State $\mu_B,\mu_G,\mu_R \gets \text{Mean}(B),\text{Mean}(G),\text{Mean}(R)$
\If{$\mu_B=0 \lor \mu_G=0 \lor \mu_R=0$}
    \State keep sample, skip normalization, log occurrence \Comment{Eq~\ref{eq:graycorld} degenerate case}
\Else
    \State $\mu \gets (\mu_B+\mu_G+\mu_R)/3$
    \State $B',G',R' \gets \text{clip}(B\mu/\mu_B,0,255),\ \text{clip}(G\mu/\mu_G,0,255),\ \text{clip}(R\mu/\mu_R,0,255)$
\EndIf
\State $I' \gets \text{Merge}(B',G',R')$; $I_\text{gray}\gets\text{RGB2Gray}(I')$
\State $M_\text{cell} \gets I_\text{gray} > 10$
\State $T \gets \text{AdaptiveGaussianThreshold}(G',\ b{=}21,\ C{=}5)$
\State $S_\text{raw} \gets T \cap M_\text{cell}$
\State $S_\text{clean} \gets \text{MorphologicalOpen}(S_\text{raw},\ \text{ellipse}(3{\times}3))$
\State $\{c_i\} \gets \text{FindContours}(S_\text{clean})$; $N_\text{spots}\gets|\{c_i\}|$
\If{$N_\text{spots}>0$}
    \State $A_\text{max}\gets\max_i\text{area}(c_i)$; $A_\text{total}\gets\sum_i\text{area}(c_i)$
    \State $c_\text{max}\gets\arg\max_i\text{area}(c_i)$; $S_\text{max}\gets\text{mean}(I'_\text{HSV,sat}[M_{c_\text{max}}])$
\Else
    \State $A_\text{max},A_\text{total},S_\text{max}\gets 0,0,0$
\EndIf
\State $\sigma_\text{gray}\gets\text{StdDev}(I_\text{gray}[M_\text{cell}])$
\State \Return $\mathbf{F}=[N_\text{spots},A_\text{max},A_\text{total},S_\text{max},\sigma_\text{gray}]$
\end{algorithmic}
\end{algorithm}

\begin{figure}[!h]
\centering
\includegraphics[width=0.7\linewidth]{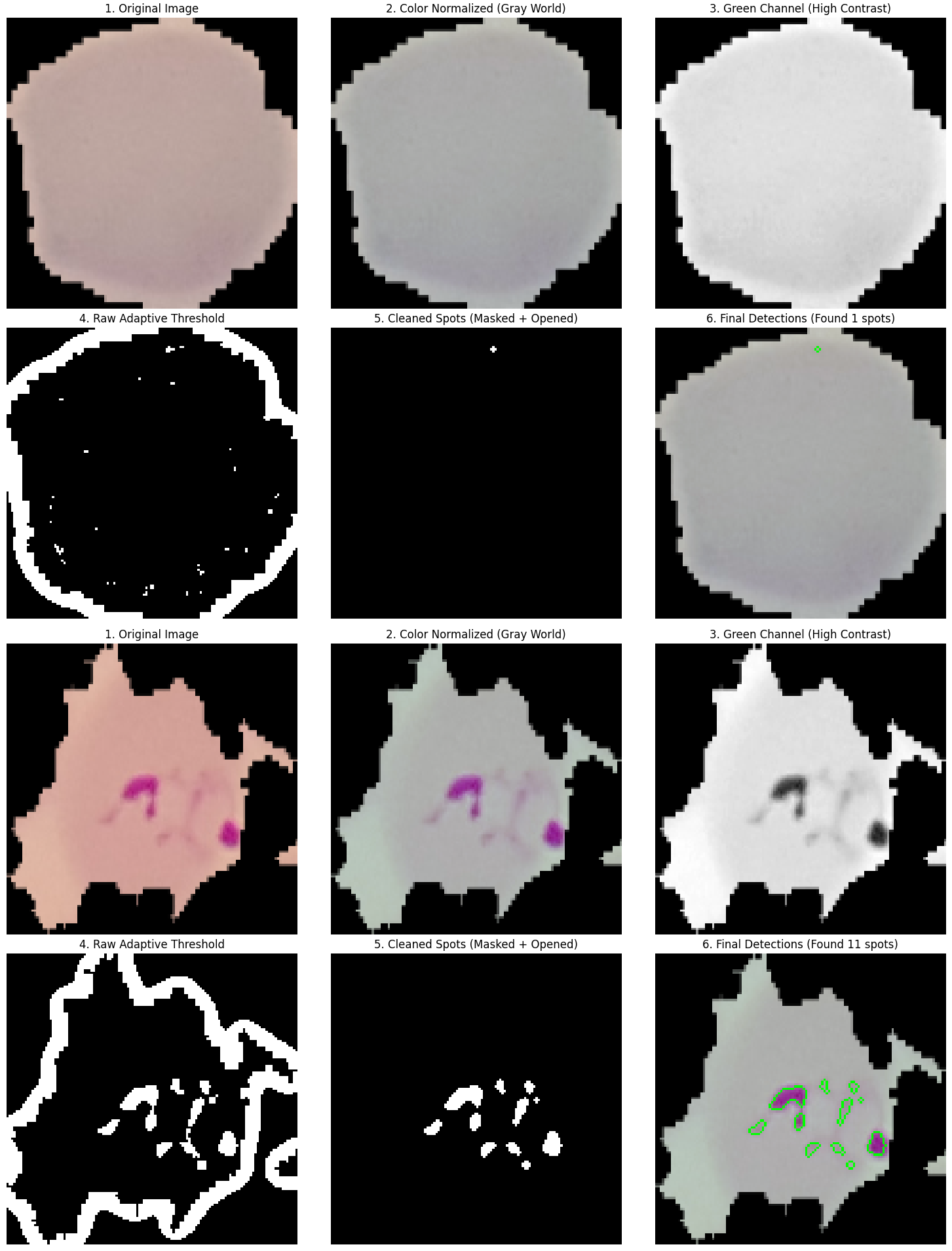}
\caption{\textbf{Comparative visualization of the feature-extraction process.} Original image, color normalization, isolated green channel, raw adaptive threshold, morphologically cleaned spot mask, and final detected contours, for an uninfected cell (top) and a parasitized cell (bottom).}
\label{fig:pipeline}
\end{figure}

Fig~\ref{fig:pipeline} shows the pipeline applied to a representative uninfected and parasitized cell, illustrating each intermediate stage.

\subsection{Classifiers and hyperparameter optimization}

Three classical classifiers are evaluated: Support Vector Machine (SVM, radial-basis or linear kernel), Random Forest~\cite{Breiman2001}, and Histogram Gradient Boosting. Hyperparameters for each are selected by Bayesian optimization, which builds a probabilistic surrogate model of the objective (validation accuracy as a function of hyperparameters) and uses it to select the next candidate configuration to evaluate, rather than exhaustively enumerating a grid or sampling uniformly at random. This is more sample-efficient than grid or random search when each evaluation is expensive (here, a full model fit), letting the search concentrate on promising regions of the hyperparameter space within a fixed evaluation budget of $n_\text{iter}=15$ candidates per fold. Table~\ref{tab:searchspace} gives the search space for each classifier.

\begin{table}[!ht]
\centering
\caption{\textbf{Bayesian hyperparameter search spaces.}}
\begin{tabular}{|p{1.2in}|p{3.6in}|}
\hline
\textbf{Classifier} & \textbf{Search space} \\
\thickhline
SVM & $C \in [10^{-1}, 10^{2}]$ (log-uniform); kernel $\in \{$RBF, linear$\}$; $\gamma \in \{$scale, auto$\}$ \\
\hline
Random Forest & $n_\text{estimators} \in [50, 300]$; $\text{max\_depth} \in [3, 20]$; $\text{min\_samples\_split} \in [2, 10]$ \\
\hline
Hist.\ Gradient Boosting & learning rate $\in [0.01, 0.2]$ (log-uniform); $\text{max\_depth} \in [3, 15]$; $\text{max\_iter} \in [50, 400]$ \\
\hline
\end{tabular}
\label{tab:searchspace}
\end{table}

\subsection{Nested cross-validation protocol}
\label{sec:nestedcv}

Every cross-validation split in this study, both the outer generalization-estimation split and the inner hyperparameter-selection split, uses \texttt{StratifiedGroupKFold}, grouped on patient identity (Section~\ref{sec:methods}) and stratified by class within that grouping constraint. The evaluation is nested: an outer split with $K_\text{outer}=20$ folds estimates generalization performance, and for each outer training fold, hyperparameters are selected independently by Bayesian optimization using an inner split with $K_\text{inner}=3$ folds, itself grouped on the outer training fold's patients only. Hyperparameter selection for a given outer fold therefore never observes that fold's test patients at any stage: not for feature standardization, not for model selection, and not for final fitting. This directly distinguishes the protocol from a non-nested design, in which a single hyperparameter search is performed once over the entire dataset before outer-fold evaluation, allowing information from every eventual test fold to influence model selection indirectly.

Before any fold is fit, an explicit runtime assertion verifies that the intersection of training-fold and test-fold patient IDs is empty; this assertion executes on every fold of every analysis reported in this study (nested cross-validation, the holdout split, the ablation study, the robustness analysis, and the deep-learning baseline comparison alike) and halts execution immediately if violated, rather than being a one-off manual check. All splits use a fixed random seed (42). Out-of-fold predictions from the 20 outer folds are pooled into a single set of predictions covering every one of the 27{,}558 images exactly once, so that reported metrics reflect the full dataset rather than a single selectively-reported fold.

Table~\ref{tab:setup} consolidates the full experimental configuration.

\begin{table}[!ht]
\centering
\caption{\textbf{Experimental setup.}}
\begin{tabular}{|p{1.7in}|p{3.1in}|}
\hline
\textbf{Parameter} & \textbf{Value} \\
\thickhline
Dataset & NIH LHNCBC, 27{,}558 images, 200 patients (150 infected / 50 control) \\
\hline
Class balance & 13{,}779 Parasitized / 13{,}779 Uninfected (exact) \\
\hline
Excluded images & 0 \\
\hline
Outer folds ($K_\text{outer}$) & 20, \texttt{StratifiedGroupKFold}, patient-grouped \\
\hline
Inner folds ($K_\text{inner}$) & 3, \texttt{StratifiedGroupKFold}, patient-grouped \\
\hline
Hyperparameter search & Bayesian optimization, $n_\text{iter}=15$ per outer fold \\
\hline
Optimization objective & Accuracy (inner cross-validation) \\
\hline
Random seed & 42 (all splits) \\
\hline
Leakage guard & Runtime assertion, every fold, every analysis \\
\hline
Feature standardization & \texttt{StandardScaler}, fit within each training fold only \\
\hline
Software & Python 3.12.10, scikit-learn, scikit-optimize \\
\hline
\end{tabular}
\label{tab:setup}
\end{table}

\subsection{Statistical validation methodology}
\label{sec:stats}

For every classifier, we report accuracy, balanced accuracy, sensitivity, specificity, positive and negative predictive value, F1 score, ROC-AUC, precision-recall AUC, and Brier score, with \textit{Parasitized} as the explicitly defined positive class. Ninety-five percent confidence intervals are computed across the 20 per-fold metric values using a $t$-interval; because folds are patient-disjoint by construction, this is a meaningful improvement over an interval computed on an image-level split, though it treats the fold, not the patient, as the unit of resampling. The held-out patient-level test set (Section~\ref{sec:holdout}) additionally reports a patient-cluster-bootstrap interval, which resamples at the patient level directly.

Statistical significance against a chance baseline is assessed with a patient-level permutation test. A one-sample $t$-test is not appropriate here, because cross-validation fold accuracies from overlapping training data are not independent observations, violating the independence assumption the test requires. Because the official NIH patient mapping confirms that all 150 infected patients contribute cells to \textit{both} classes (an infected patient's slide is a mixture of parasitized and uninfected-looking cells), a patient-level label swap (assigning each patient a single shuffled label) is not a valid null-generation strategy here. Instead, cell labels are shuffled globally while evaluation continues to run through the same patient-grouped cross-validation structure, testing whether the patient-grouped protocol itself would still report high accuracy on pure noise labels. For tractability at 1{,}000 permutations, the classifier's hyperparameters are frozen at the modal configuration selected across the 20 outer folds of the main evaluation, and each permutation is evaluated through a fresh 5-fold (not 20-fold) patient-grouped cross-validation, with the leakage-guard assertion enforced on every one of these folds as well. The $p$-value is computed as
\begin{equation}
\label{eq:pvalue}
p = \frac{1 + \#\{\text{null} \geq \text{observed}\}}{n_\text{permutations} + 1}.
\end{equation}

Comparisons between two models evaluated on the identical set of cross-validation folds (the ablation study and the deep-learning baseline comparison, both below) use a paired Wilcoxon signed-rank test on matched per-fold metric values, following the standard recommendation for comparing classifiers across cross-validation folds~\cite{Demsar2006}, rather than an unpaired test or a bare point-estimate comparison.

\subsection{Held-out patient-level test set}
\label{sec:holdout}

As a complement to the pooled 20-fold estimate, we additionally evaluate a single, genuinely untouched patient-level train/test split, obtained as one fold of a 5-way \texttt{StratifiedGroupKFold} partition (160 patients / 21{,}915 cells for training and hyperparameter selection; 40 patients / 5{,}643 cells held out entirely from model selection). Hyperparameters are selected via Bayesian optimization using only the training partition (with its own patient-grouped inner cross-validation), so the 40 held-out patients influence neither feature standardization, hyperparameter selection, nor training; they are scored exactly once. A 95\% confidence interval is obtained via a patient-cluster bootstrap: the \textit{set of test patients} (not individual cells) is resampled with replacement 2{,}000 times, with every cell belonging to a resampled patient carried along in that resample, and the metric of interest recomputed on each resample.

\subsection{Ablation study design}
\label{sec:ablationdesign}

To evaluate whether the five-feature design outperforms simpler alternatives and to isolate each component's individual contribution, we compare against four classical baseline descriptors, namely color histograms~\cite{Swain1991}, Local Binary Patterns~\cite{Ojala2002}, Gray-Level Co-occurrence Matrix texture features~\cite{Haralick1973}, and Histogram of Oriented Gradients~\cite{Dalal2005}, substituted in place of the EMFE feature vector and evaluated with Random Forest under the identical patient-grouped protocol. We additionally evaluate four stage ablations (disabling color normalization; substituting a single global Otsu threshold or a fixed intensity threshold for the adaptive threshold; disabling morphological opening) and a leave-one-feature-out sweep (each of the five features removed in turn), each compared to the full pipeline via the paired Wilcoxon test described above. For computational tractability, this comparison uses a reduced protocol ($K_\text{outer}=5$, $K_\text{inner}=2$, $n_\text{iter}=8$) rather than the full 20-fold protocol. With only 5 paired folds, the smallest attainable two-sided Wilcoxon $p$-value is $0.0625$ regardless of effect size; entries at that value reflect this resolution limit rather than a specific significance level.

\subsection{Robustness analysis design}
\label{sec:robustnessdesign}

To probe sensitivity to imaging and staining variation without access to an externally-sourced dataset, we apply seven synthetic perturbations, each at two severity levels, to the held-out test set (Section~\ref{sec:holdout}): brightness scaling ($\times 0.7$ / $\times 1.3$), contrast scaling ($\times 0.7$ / $\times 1.3$), hue shift ($10^\circ$ / $20^\circ$), additive Gaussian noise ($\sigma=5$ / $\sigma=15$), JPEG re-compression ($q=30$ / $q=10$), Gaussian blur ($\sigma=1$ / $\sigma=3$), and a downscale-upscale resolution round-trip ($64$px / $32$px intermediate resolution). Each perturbed image is passed through the complete, unmodified pipeline (Section~\ref{sec:pipeline}), including its own color normalization, and scored with the frozen Random Forest holdout model (Section~\ref{sec:holdout}), with no retraining or perturbation-specific adaptation. We emphasize that a synthetic perturbation study of this kind provides an additional robustness signal but does not substitute for evaluation on externally-sourced data (different microscopes, cameras, staining batches, or laboratories); it is reported and interpreted accordingly.

\subsection{Deep-learning baseline comparison design}
\label{sec:dldesign}

We compare EMFE against three convolutional architectures pretrained on ImageNet and fully fine-tuned end to end (no frozen layers): DenseNet121 and ResNet50, two widely used large general-purpose architectures, and MobileNetV2, a lightweight architecture designed for mobile and edge deployment, included specifically to test whether the accuracy/efficiency trade-off characterized in this study also holds against an efficiency-oriented deep-learning alternative, not only against large architectures. All three are trained and evaluated under the same patient-grouped protocol, the same $K_\text{outer}=20$ outer folds, and the same source images as EMFE's own headline evaluation (Section~\ref{sec:nestedcv}). Sharing the identical fold partition is what makes the paired Wilcoxon comparison in the Results section valid without a protocol-mismatch caveat. Images are resized to $224\times224$ (each architecture's native ImageNet input resolution) rather than EMFE's own $128\times128$. Each architecture is fine-tuned for a fixed 6 epochs per fold (Adam optimizer, learning rate $10^{-4}$, batch size 64, automatic mixed precision) on a single GPU; epoch count, learning rate, and batch size are fixed instead of hyperparameter-searched, disclosed explicitly as a tractability trade-off given the cost of a full architecture $\times$ fold sweep, consistent with the same fixed-protocol trade-off already used for the ablation study above. Model size is the actual serialized parameter state, measured directly and not assumed from published architecture specifications. Inference latency is measured on both CPU and GPU, from a single trained checkpoint per architecture, using the same staged (decode / preprocess / predict measured separately) timing methodology as the classical pipeline's own benchmark (Section~\ref{sec:effdesign}), so that CPU and GPU figures for a given architecture come from one controlled measurement rather than being pooled from separate runs.

\subsection{Explainability methodology}
\label{sec:explaindesign}

Using the frozen Random Forest holdout model and its held-out test set (Section~\ref{sec:holdout}), we report: (i) permutation feature importance, computed as the mean decrease in accuracy over 30 repeats of randomly permuting each feature column independently, with the resulting repeat-to-repeat standard deviation reported as an explicit uncertainty estimate; (ii) partial dependence of the predicted \textit{Parasitized} probability on each feature individually, with the remaining four features held at their observed distribution; (iii) a labeled gallery of representative true-positive, true-negative, false-positive, and false-negative examples; and (iv) a quantitative failure-mode comparison, testing whether each feature's distribution differs significantly between correct and incorrect predictions within each true class (false negatives vs.\ true positives; false positives vs.\ true negatives) via a two-sided Mann-Whitney $U$ test. We additionally report a one-at-a-time sensitivity sweep over the three adaptive-thresholding parameters (block size, $C$, morphological kernel size) against the pipeline's default configuration, under the same reduced 5-fold protocol and paired Wilcoxon comparison as the ablation study.

\subsection{Patient-level aggregation methodology}
\label{sec:patientleveldesign}

EMFE itself produces a per-cell prediction; it does not implement any rule for aggregating many single-cell predictions into a patient-level call. To characterize what such a rule's behavior would look like, we perform a post hoc analysis using the patient-disjoint out-of-fold predictions already produced by the main nested cross-validation (Section~\ref{sec:nestedcv}). Every cell's prediction came from an outer fold whose training set never included that cell's patient, so this analysis is as leakage-free at the patient level as the underlying per-cell result. Predictions are grouped by all 200 distinct patients in the dataset (150 infected, 50 uninfected-only control, using the NIH dataset's own cohort structure as ground truth rather than an invented threshold), and the rule ``call a patient infected if at least $\tau$ of their cells are predicted \textit{Parasitized}'' is evaluated by sweeping $\tau \in \{1, 2, 3, 5, 10, 20\}$. Patient-level sensitivity is the fraction of the 150 infected patients flagged; specificity is the fraction of the 50 control patients \textit{not} flagged. Ninety-five percent confidence intervals use the Wilson score interval, appropriate given some strata are as small as 37--50 patients and observed proportions can sit at or near 0 or 1, where a normal approximation is unreliable. We additionally quantify false-positive accumulation directly: given a per-cell false-positive rate $r$ and a control patient with $n$ cells, simple compounding predicts $P(\text{at least one false positive}) = 1-(1-r)^n$; we compare this prediction, averaged over the actual per-control-patient cell counts, against the observed flag rate at $\tau=1$. Sensitivity is additionally stratified by true parasitemia (the fraction of a patient's cells that are truly parasitized), using data-driven quartile boundaries of the observed parasitemia distribution across the 150 infected patients, to test whether patients with a low true parasite burden are disproportionately missed.

\subsection{Computational efficiency benchmarking protocol}
\label{sec:effdesign}

Timing, memory, and CPU-utilization figures for the classical pipeline are measured over 500 timed repetitions (following 50 discarded warm-up repetitions) cycling through real dataset images on a development workstation (Intel Core i5-13400; workload pinned to its 6 performance cores only, 12 logical threads, with the 4 efficiency cores disabled; CPU power limit fixed at 35~W via BIOS PL1=PL2; Windows 11, Python 3.12.10). All hardware and software conditions were held fixed across repetitions; ordinary OS-level background scheduling is the one residual, uncontrolled source of run-to-run timing variance. Image decode, feature extraction, and prediction are timed as three separate stages rather than a single pooled figure, so that each reported number states precisely what it includes. Model size is the actual compressed on-disk artifact size (\texttt{joblib} compression level 3), reported separately from the resident-memory delta observed after loading that artifact and the peak resident memory observed during the timed inference loop; these are three distinct quantities that are frequently conflated in this literature. CPU utilization is reported as a raw system-wide reading alongside the exact core count it was measured against.

\subsection{Ethics statement}

This study uses only the publicly available, de-identified NIH LHNCBC malaria cell-image dataset~\cite{MalariaProject}, which contains no personally identifiable information. No new human-subjects data were collected. As the dataset is fully de-identified and publicly released for research use by its original custodians, no additional institutional ethics review was required for its secondary use in this study.

\section{Results}
\label{sec:results}

\subsection{Predictive performance}
\label{sec:res-predictive}

Table~\ref{tab:mainresults} reports the complete diagnostic-metrics suite for all three classifiers under the patient-grouped nested cross-validation protocol (Section~\ref{sec:nestedcv}), pooled across all 27{,}558 out-of-fold predictions. All three classifiers converge to statistically indistinguishable accuracy (94.6\%). Random Forest is used as the primary model throughout the remainder of this section: its trees are trained and evaluated independently of one another, so both training and inference parallelize across CPU cores in a way that Histogram Gradient Boosting's sequential tree construction and SVM's single-threaded optimization do not, making it the more practical default given this study's CPU-only deployment focus. Histogram Gradient Boosting and SVM are reported alongside it throughout for completeness, since accuracy alone does not distinguish between them.

\begin{adjustwidth}{-.75in}{-.75in}
\centering
\begin{table}[!ht]
\centering
\caption{\textbf{Pooled out-of-fold metrics under patient-grouped nested cross-validation} ($n=27{,}558$; 95\% CI across the 20 outer folds).}
\begin{tabular}{|l|c|c|c|}
\hline
\textbf{Metric} & \textbf{Random Forest} & \textbf{Hist.\ Gradient Boosting} & \textbf{SVM} \\
\thickhline
Accuracy & 0.9463 [0.9360, 0.9572] & 0.9462 [0.9362, 0.9568] & 0.9463 [0.9358, 0.9579] \\
\hline
Balanced accuracy & 0.9463 [0.9358, 0.9564] & 0.9462 [0.9356, 0.9561] & 0.9463 [0.9354, 0.9569] \\
\hline
Sensitivity & 0.9335 [0.9169, 0.9495] & 0.9316 [0.9147, 0.9472] & 0.9320 [0.9149, 0.9484] \\
\hline
Specificity & 0.9590 [0.9498, 0.9682] & 0.9607 [0.9521, 0.9694] & 0.9607 [0.9509, 0.9704] \\
\hline
PPV & 0.9579 [0.9277, 0.9674] & 0.9596 [0.9296, 0.9685] & 0.9595 [0.9301, 0.9690] \\
\hline
NPV & 0.9352 [0.9167, 0.9572] & 0.9336 [0.9153, 0.9552] & 0.9339 [0.9149, 0.9566] \\
\hline
F1 & 0.9456 [0.9248, 0.9547] & 0.9454 [0.9244, 0.9544] & 0.9456 [0.9249, 0.9551] \\
\hline
ROC-AUC & 0.9765 [0.9664, 0.9839] & 0.9768 [0.9674, 0.9852] & 0.9670 [0.9588, 0.9776] \\
\hline
PR-AUC & 0.9809 [0.9637, 0.9861] & 0.9816 [0.9650, 0.9871] & 0.9729 [0.9594, 0.9784] \\
\hline
Brier score & 0.0440 & 0.0432 & 0.0451 \\
\hline
\end{tabular}
\label{tab:mainresults}
\end{table}
\end{adjustwidth}

\begin{figure}[!h]
\centering
\includegraphics[width=0.55\linewidth]{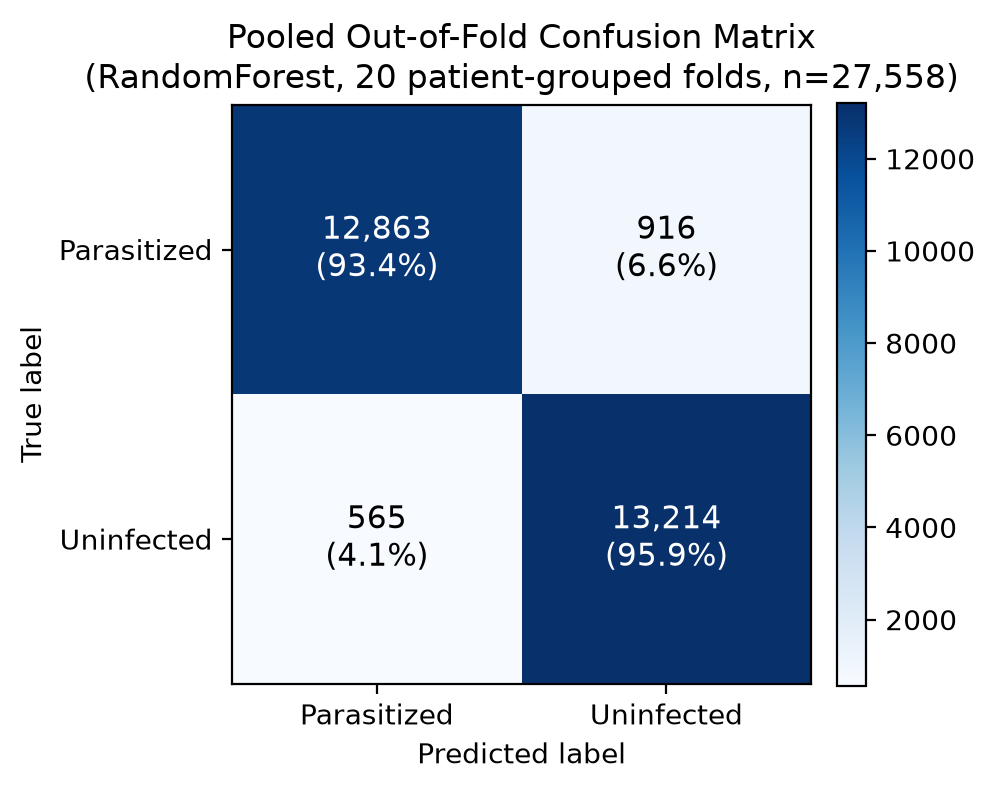}
\caption{\textbf{Pooled out-of-fold confusion matrix.} Random Forest predictions across all 20 patient-grouped outer folds ($n=27{,}558$), with counts and row-wise percentages for each true class.}
\label{fig:confusion}
\end{figure}

\begin{figure}[!h]
\centering
\includegraphics[width=\linewidth]{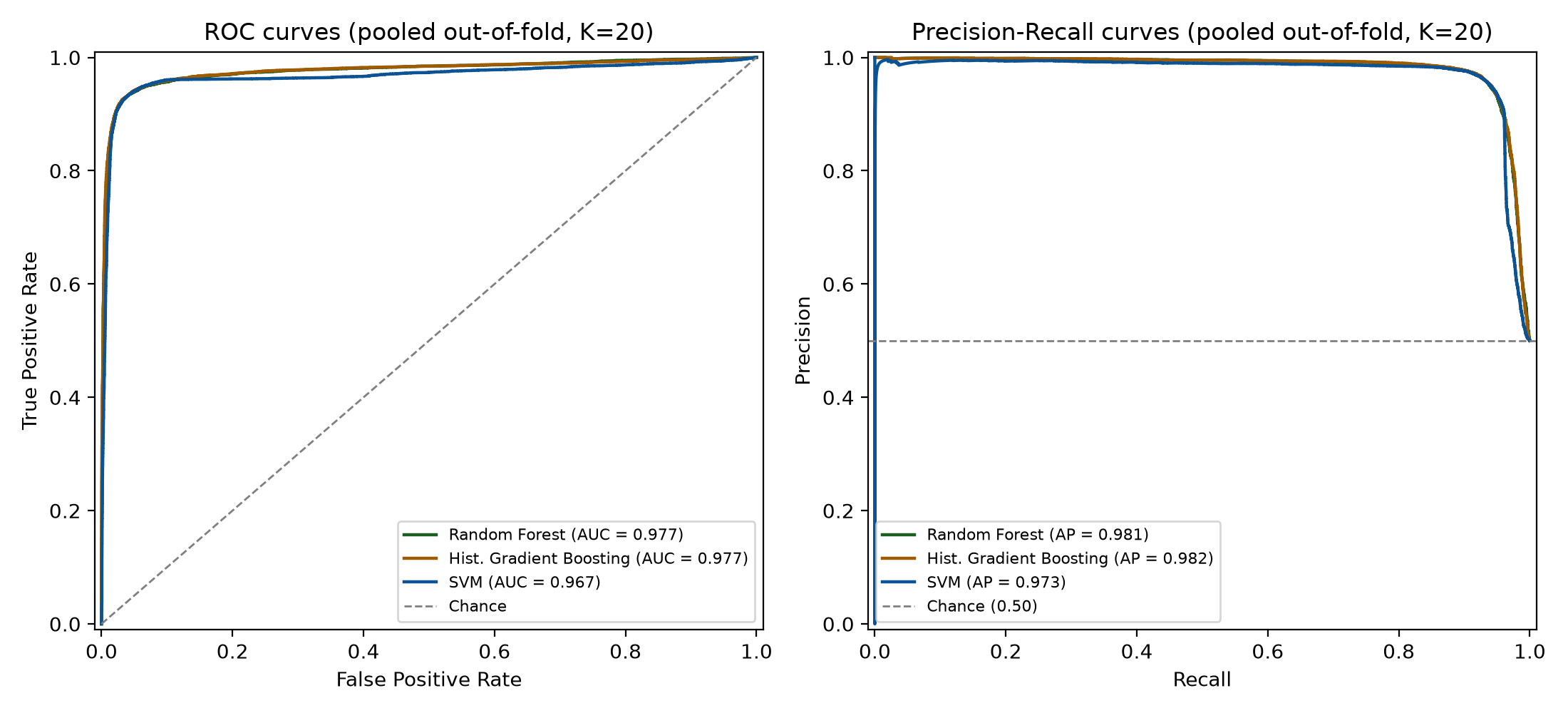}
\caption{\textbf{ROC and precision-recall curves.} Pooled out-of-fold ROC curves (left) and precision-recall curves (right) for all three classifiers, computed from the 27{,}558 pooled out-of-fold predictions of the patient-grouped nested cross-validation. AUC and average precision (AP) values match Table~\ref{tab:mainresults}'s ROC-AUC and PR-AUC columns exactly, as both are computed from the identical prediction set.}
\label{fig:roc}
\end{figure}

Of the 13{,}779 truly parasitized cells, 12{,}863 (93.4\%) were correctly classified by the Random Forest model and 916 (6.6\%) were missed; of the 13{,}779 truly uninfected cells, 13{,}214 (95.9\%) were correctly classified and 565 (4.1\%) were false alarms. Fig~\ref{fig:confusion} shows the corresponding pooled out-of-fold confusion matrix, and Fig~\ref{fig:roc} shows the corresponding ROC and precision-recall curves for all three classifiers, computed directly from these pooled out-of-fold prediction scores.

\subsection{Statistical significance}

The patient-level permutation test (Section~\ref{sec:stats}) gives an observed accuracy of 0.9461 against a null distribution with mean 0.4998 and standard deviation 0.0038, consistent with chance, as expected under global label shuffling. The observed accuracy exceeded all 1{,}000 permuted-label runs, giving $p = (1+0)/(1000+1) \approx 0.000999$, reported as $p<0.001$ (the resolution limit imposed by 1{,}000 permutations).

\subsection{Held-out patient-level test set}
\label{sec:res-holdout}

Table~\ref{tab:holdout} reports accuracy on the untouched 40-patient (5{,}643-cell) holdout partition (Section~\ref{sec:holdout}), with a 95\% patient-cluster-bootstrap confidence interval. All three models score within 0.4 percentage points of one another and within 0.4 points of the pooled 20-fold estimate (Table~\ref{tab:mainresults}), on patients never used for model selection, corroborating that estimate.

\begin{table}[!ht]
\centering
\caption{\textbf{Held-out patient-level test set} (160 train / 40 test patients; 21{,}915 / 5{,}643 cells; 95\% patient-cluster-bootstrap CI).}
\begin{tabular}{|l|c|c|c|}
\hline
\textbf{Metric} & \textbf{Random Forest} & \textbf{Hist.\ Gradient Boosting} & \textbf{SVM} \\
\thickhline
Accuracy & 0.9429 [0.9263, 0.9582] & 0.9424 [0.9255, 0.9581] & 0.9440 [0.9277, 0.9591] \\
\hline
\end{tabular}
\label{tab:holdout}
\end{table}

\subsection{Ablation study}
\label{sec:res-ablation}

Table~\ref{tab:baseline} compares the five-feature EMFE pipeline against four classical baseline descriptors, all evaluated with Random Forest under the reduced protocol described in Section~\ref{sec:ablationdesign}. The EMFE pipeline outperforms every baseline, most by a wide margin; the general-purpose texture descriptors (LBP, GLCM) trail by 23--29 accuracy points, well below the purpose-built spot-detection approach.

\begin{table}[!ht]
\centering
\caption{\textbf{Baseline-descriptor comparison} (Random Forest, reduced 5-fold protocol) against the full EMFE pipeline (accuracy 0.9460, F1 0.9453).}
\begin{tabular}{|l|c|c|c|}
\hline
\textbf{Descriptor} & \textbf{Accuracy} & \textbf{F1} & $\boldsymbol{\Delta}$ \textbf{Accuracy} \\
\thickhline
Color histogram~\cite{Swain1991} & 0.9311 & 0.9299 & $-0.0149$ \\
\hline
Local Binary Patterns~\cite{Ojala2002} & 0.7061 & 0.7078 & $-0.2284$ \\
\hline
Gray-Level Co-occurrence Matrix~\cite{Haralick1973} & 0.6448 & 0.6516 & $-0.2917$ \\
\hline
Histogram of Oriented Gradients~\cite{Dalal2005} & 0.8567 & 0.8591 & $-0.0832$ \\
\hline
\end{tabular}
\label{tab:baseline}
\end{table}

Table~\ref{tab:fullablation} reports the four stage ablations and the leave-one-feature-out sweep. With only 5 paired folds, the smallest achievable two-sided Wilcoxon $p$-value is 0.0625, a resolution limit rather than a measure of effect size; where entries reach this floor, all 5 folds nonetheless changed in the same direction.

\begin{adjustwidth}{-.5in}{-.5in}
\centering
\begin{table}[!ht]
\centering
\caption{\textbf{Stage-ablation and leave-one-feature-out results} (Random Forest, reduced 5-fold protocol, default configuration: accuracy 0.9463, F1 0.9456).}
\begin{tabular}{|l|c|c|c|}
\hline
\textbf{Variant} & \textbf{Accuracy} & \textbf{F1} & \textbf{Wilcoxon} $\boldsymbol{p}$ \\
\thickhline
\multicolumn{4}{|l|}{\textit{Stage ablations}} \\
\hline
No color normalization & 0.9435 & 0.9431 & 0.4375 \\
\hline
Otsu threshold (vs.\ adaptive) & 0.9381 & 0.9351 & 0.1875 \\
\hline
Fixed global threshold & 0.9430 & 0.9425 & 1.0000 \\
\hline
No morphological opening & 0.9430 & 0.9421 & 0.0625 \\
\hline
\multicolumn{4}{|l|}{\textit{Leave-one-feature-out}} \\
\hline
Without $N_\text{spots}$ & 0.9420 & 0.9411 & 0.0625 \\
\hline
Without $A_\text{max}$ & 0.9451 & 0.9443 & 0.4375 \\
\hline
Without $A_\text{total}$ & 0.9408 & 0.9399 & 0.0625 \\
\hline
Without $S_\text{max}$ (saturation) & 0.9200 & 0.9211 & 0.0625 \\
\hline
Without $\sigma_\text{gray}$ & 0.9468 & 0.9460 & 0.1875 \\
\hline
\end{tabular}
\label{tab:fullablation}
\end{table}
\end{adjustwidth}

Removing spot saturation ($S_\text{max}$) produces the largest single-feature accuracy drop (2.6 points), at the maximum-resolution significance level for this protocol, providing direct empirical confirmation of the biological rationale given in Section~\ref{sec:pipeline}. Removing $\sigma_\text{gray}$ instead slightly improves accuracy, suggesting it contributes little beyond what the other four features already capture. None of the four stage ablations reach significance at this fold count, though disabling color normalization and switching to a global Otsu threshold both trend toward lower accuracy.

\subsection{Robustness to synthetic imaging perturbations}
\label{sec:res-robustness}

Table~\ref{tab:robustness} reports the full synthetic-perturbation sweep against the unperturbed holdout baseline (accuracy 0.9429, F1 0.9429). The pipeline is largely robust to brightness, hue-shift, and moderate noise/compression changes (within roughly 2--4 points of baseline in most cases). It is not robust to three specific perturbations: reduced contrast, strong Gaussian blur, and aggressive resolution downscaling, each of which collapses accuracy to near chance.

\begin{adjustwidth}{-.75in}{-.75in}
\centering
\begin{table}[!ht]
\centering
\caption{\textbf{Full synthetic-perturbation robustness sweep.}}
\begin{tabular}{|l|c|c|c|c|}
\hline
\textbf{Perturbation} & \textbf{Accuracy} & \textbf{F1} & \textbf{Sensitivity} & \textbf{Specificity} \\
\thickhline
Brightness, dimmer ($\times0.7$) & 0.9415 & 0.9409 & 0.9068 & 0.9782 \\
\hline
Brightness, brighter ($\times1.3$) & 0.9240 & 0.9248 & 0.9106 & 0.9381 \\
\hline
Contrast, reduced ($\times0.7$) & 0.5150 & 0.6791 & 1.0000 & 0.0036 \\
\hline
Contrast, increased ($\times1.3$) & 0.9279 & 0.9286 & 0.9137 & 0.9428 \\
\hline
Hue shift, $10^\circ$ & 0.9421 & 0.9425 & 0.9254 & 0.9596 \\
\hline
Hue shift, $20^\circ$ & 0.9454 & 0.9459 & 0.9302 & 0.9614 \\
\hline
Gaussian noise, $\sigma=5$ & 0.9444 & 0.9441 & 0.9154 & 0.9749 \\
\hline
Gaussian noise, $\sigma=15$ & 0.9243 & 0.9221 & 0.8729 & 0.9785 \\
\hline
JPEG compression, $q=30$ & 0.8570 & 0.8646 & 0.8898 & 0.8224 \\
\hline
JPEG compression, $q=10$ & 0.8988 & 0.8987 & 0.8743 & 0.9246 \\
\hline
Gaussian blur, $\sigma=1$ & 0.9226 & 0.9229 & 0.9030 & 0.9432 \\
\hline
Gaussian blur, $\sigma=3$ & 0.5504 & 0.6849 & 0.9520 & 0.1270 \\
\hline
Resolution downscale, 64px & 0.9281 & 0.9271 & 0.8919 & 0.9661 \\
\hline
Resolution downscale, 32px & 0.5860 & 0.7063 & 0.9700 & 0.1813 \\
\hline
\end{tabular}
\label{tab:robustness}
\end{table}
\end{adjustwidth}

In each of the three failure cases, sensitivity remains high (0.91--1.00) while specificity collapses (0.004--0.18): the model defaults to predicting \textit{Parasitized} for nearly every cell once the perturbation disrupts spot detection, rather than failing randomly across both classes. This is mechanistically consistent with the pipeline's dependence on local intensity contrast (Section~\ref{sec:pipeline}): reduced global contrast, blur, and low resolution each directly destroy the contrast between candidate parasite regions and their surroundings that adaptive thresholding depends on, producing spurious spot detections across the whole cell rather than a clean absence of signal.

\subsection{Deep-learning baseline comparison}
\label{sec:res-dl}

Table~\ref{tab:dlbaselines} compares EMFE against DenseNet121, ResNet50, and MobileNetV2, all retrained and evaluated under the identical patient-grouped protocol, identical 20 outer folds, and identical source images as EMFE's own headline evaluation (Section~\ref{sec:dldesign}).

\begin{adjustwidth}{-.75in}{-.75in}
\centering
\begin{table}[!ht]
\centering
\caption{\textbf{Deep-learning baselines}, retrained under the identical patient-grouped protocol as EMFE's headline evaluation (pooled over 20 outer folds; $n=27{,}558$).}
\begin{tabular}{|l|c|c|c|c|c|}
\hline
\textbf{Model} & \textbf{Accuracy} & \textbf{F1} & \textbf{Size} & \textbf{Latency (CPU)} & \textbf{Latency (GPU)} \\
\thickhline
EMFE (Random Forest) & 0.9463 & 0.9456 & 6.3 MB & 4.6 ms & N/A (CPU-only) \\
\hline
DenseNet121 & 0.9698 & 0.9695 & 27.1 MB & 197.4 ms & 35.7 ms \\
\hline
ResNet50 & 0.9676 & 0.9674 & 90.0 MB & 105.4 ms & 11.7 ms \\
\hline
MobileNetV2 & 0.9670 & 0.9666 & 8.7 MB & 17.3 ms & 4.8 ms \\
\hline
\end{tabular}
\label{tab:dlbaselines}
\end{table}
\end{adjustwidth}

On identical CPU hardware, EMFE is 3.8--43$\times$ faster than the three deep baselines (3.8$\times$ against MobileNetV2, the lightest; 43$\times$ against DenseNet121, the heaviest) and 1.4--14.3$\times$ smaller on-disk, in exchange for 1.7--2.4 accuracy points. This difference is statistically significant and well-powered for every comparison: a paired Wilcoxon signed-rank test on matched per-fold accuracy across all 20 outer folds gives $p\approx1.9\times10^{-6}$ against DenseNet121, ResNet50, and MobileNetV2 alike (identical result for F1), with every deep model scoring higher than EMFE on every one of the 20 folds (Table~\ref{tab:dlwilcoxon}). Because this comparison shares EMFE's exact 20-fold partition, the accuracy figures are directly comparable without a protocol-mismatch caveat.

\begin{table}[!ht]
\centering
\caption{\textbf{Paired Wilcoxon signed-rank test}: each deep baseline vs.\ EMFE, matched per fold (20 folds, identical partition).}
\begin{tabular}{|l|c|c|c|c|}
\hline
\textbf{Comparison} & \textbf{Metric} & \textbf{Median} $\boldsymbol{\Delta}$ & \textbf{Folds won by DL model} & \textbf{Wilcoxon} $\boldsymbol{p}$ \\
\thickhline
DenseNet121 vs.\ EMFE & Accuracy & $+0.0198$ & 20 / 20 & $1.9\times10^{-6}$ \\
\hline
DenseNet121 vs.\ EMFE & F1 & $+0.0189$ & 20 / 20 & $1.9\times10^{-6}$ \\
\hline
ResNet50 vs.\ EMFE & Accuracy & $+0.0171$ & 20 / 20 & $1.9\times10^{-6}$ \\
\hline
ResNet50 vs.\ EMFE & F1 & $+0.0178$ & 20 / 20 & $1.9\times10^{-6}$ \\
\hline
MobileNetV2 vs.\ EMFE & Accuracy & $+0.0177$ & 20 / 20 & $1.9\times10^{-6}$ \\
\hline
MobileNetV2 vs.\ EMFE & F1 & $+0.0169$ & 20 / 20 & $1.9\times10^{-6}$ \\
\hline
\end{tabular}
\label{tab:dlwilcoxon}
\end{table}

\subsection{Explainability}
\label{sec:res-explain}

\begin{figure}[!h]
\centering
\includegraphics[width=\linewidth]{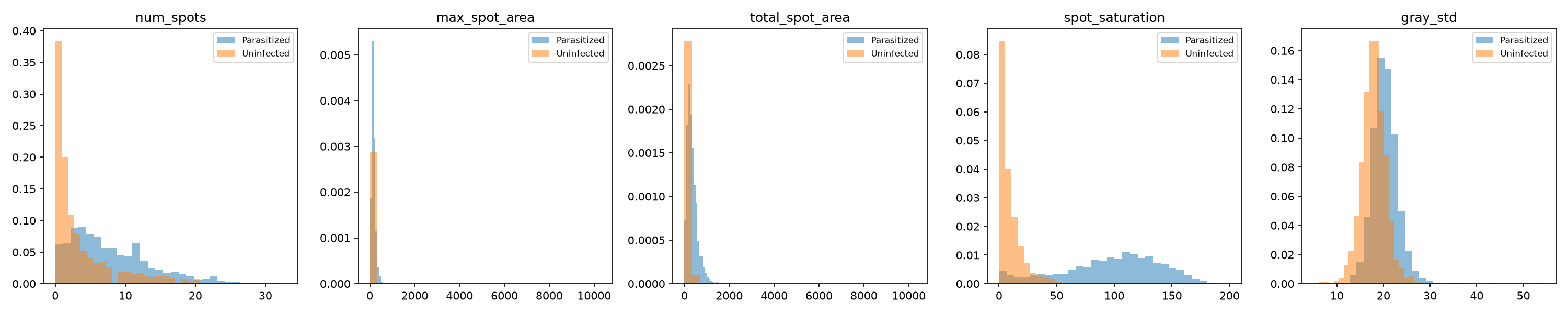}
\caption{\textbf{Per-feature distributions by class.} Distribution of each of the five EMFE features, split by true class, on the held-out test set.}
\label{fig:featdist}
\end{figure}

Fig~\ref{fig:featdist} shows the per-feature distributions by true class on the held-out test set. Table~\ref{tab:importance} reports permutation feature importance (30 repeats), corroborating spot saturation as by far the dominant feature: more than $7\times$ the importance of the next-ranked feature.

\begin{table}[!ht]
\centering
\caption{\textbf{Permutation feature importance} (30 repeats, mean $\pm$ std, held-out test set).}
\begin{tabular}{|l|c|}
\hline
\textbf{Feature} & \textbf{Importance} \\
\thickhline
Spot saturation ($S_\text{max}$) & $0.2670 \pm 0.0056$ \\
\hline
Total spot area ($A_\text{total}$) & $0.0341 \pm 0.0025$ \\
\hline
Max spot area ($A_\text{max}$) & $0.0301 \pm 0.0016$ \\
\hline
Number of spots ($N_\text{spots}$) & $0.0098 \pm 0.0010$ \\
\hline
Gray standard deviation ($\sigma_\text{gray}$) & $0.0021 \pm 0.0009$ \\
\hline
\end{tabular}
\label{tab:importance}
\end{table}

\begin{figure}[!h]
\centering
\includegraphics[width=\linewidth]{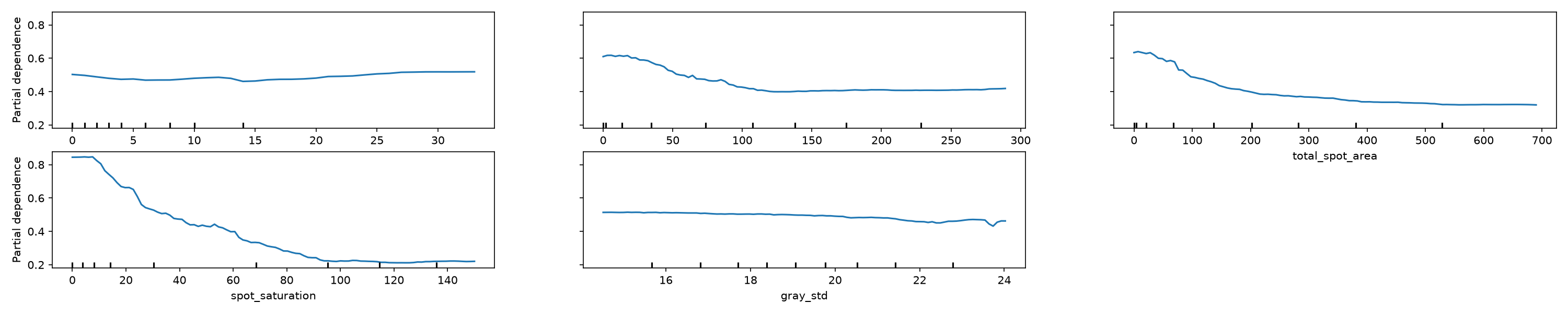}
\caption{\textbf{Partial dependence.} Partial dependence of the predicted Parasitized probability on each of the five features individually, with the other four held at their observed distribution.}
\label{fig:pdp}
\end{figure}

\begin{figure}[!h]
\centering
\includegraphics[width=0.7\linewidth]{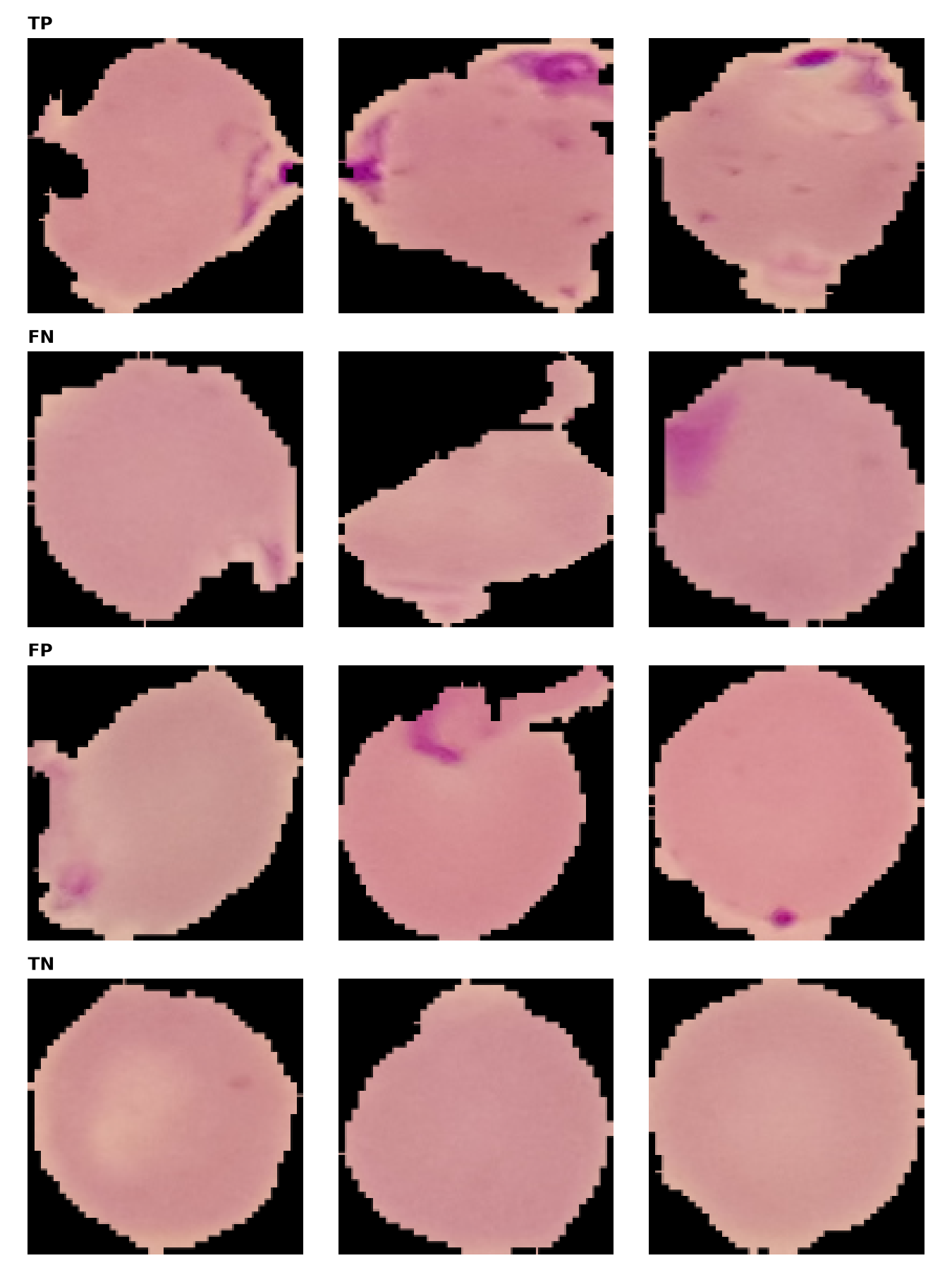}
\caption{\textbf{Representative prediction-category gallery.} True-positive, false-negative, false-positive, and true-negative example cells from the held-out test set, with their corresponding feature-extraction pipeline visualization.}
\label{fig:gallery}
\end{figure}

Fig~\ref{fig:pdp} shows the partial dependence of the predicted \textit{Parasitized} probability on each feature. Fig~\ref{fig:gallery} shows a representative labeled gallery from each of the four prediction categories (2{,}657 true positives, 239 false negatives, 2{,}664 true negatives, 83 false positives on the held-out test set).

Table~\ref{tab:failuremode} quantifies feature-value differences between correct and incorrect predictions within each true class via a two-sided Mann-Whitney $U$ test; every feature differs significantly ($p<10^{-4}$) in both comparisons.

\begin{adjustwidth}{-.75in}{-.75in}
\centering
\begin{table}[!ht]
\centering
\caption{\textbf{Failure-mode feature comparison} (Mann-Whitney $U$, held-out test set).}
\begin{tabular}{|l|c|c|c|c|c|c|}
\hline
\textbf{Feature} & \textbf{FN mean} & \textbf{TP mean} & $\boldsymbol{p}$ \textbf{(FN v.\ TP)} & \textbf{FP mean} & \textbf{TN mean} & $\boldsymbol{p}$ \textbf{(FP v.\ TN)} \\
\thickhline
$N_\text{spots}$ & 5.57 & 8.16 & $1.8\times10^{-15}$ & 6.75 & 3.35 & $9.4\times10^{-16}$ \\
\hline
$A_\text{max}$ & 72.21 & 170.87 & $2.4\times10^{-81}$ & 148.74 & 42.53 & $9.9\times10^{-44}$ \\
\hline
$A_\text{total}$ & 171.03 & 367.28 & $1.4\times10^{-52}$ & 263.85 & 70.62 & $6.4\times10^{-38}$ \\
\hline
$S_\text{max}$ & 19.06 & 104.76 & $1.2\times10^{-128}$ & 77.31 & 8.48 & $2.1\times10^{-53}$ \\
\hline
$\sigma_\text{gray}$ & 19.73 & 20.46 & $1.5\times10^{-5}$ & 20.85 & 17.77 & $2.0\times10^{-17}$ \\
\hline
\end{tabular}
\label{tab:failuremode}
\end{table}
\end{adjustwidth}

False negatives (missed parasitized cells) show systematically weaker spot signal than true positives across every feature, with fewer, smaller, and less saturated candidate regions, consistent with these being marginal, low-signal cells rather than errors unrelated to the underlying feature design. False positives (uninfected cells misclassified as parasitized) show the opposite pattern relative to true negatives: elevated spot count, area, and especially saturation, consistent with staining artifacts or debris producing a spuriously parasite-like signal.

Table~\ref{tab:threshsens} reports the threshold-parameter sensitivity sweep. The pipeline is more sensitive to the adaptive-threshold constant $C$ (accuracy range 0.892--0.954) than to block size (0.921--0.952) or morphological kernel size (0.942--0.950); the default configuration performs within 0.8 points of the best value found for either parameter.

\begin{table}[!ht]
\centering
\caption{\textbf{Threshold-parameter sensitivity} (Random Forest, reduced 5-fold protocol) against the default configuration (accuracy 0.9463, F1 0.9456).}
\begin{tabular}{|l|c|c|c|}
\hline
\textbf{Variant} & \textbf{Accuracy} & \textbf{F1} & \textbf{Wilcoxon} $\boldsymbol{p}$ \\
\thickhline
Block size 11 (default 21) & 0.9520 & 0.9513 & 0.0625 \\
\hline
Block size 31 (default 21) & 0.9312 & 0.9303 & 0.0625 \\
\hline
Block size 41 (default 21) & 0.9212 & 0.9204 & 0.0625 \\
\hline
Adaptive $C=2$ (default 5) & 0.8916 & 0.8897 & 0.0625 \\
\hline
Adaptive $C=8$ (default 5) & 0.9514 & 0.9508 & 0.0625 \\
\hline
Adaptive $C=12$ (default 5) & 0.9542 & 0.9536 & 0.0625 \\
\hline
Morph.\ kernel 1 (default 3) & 0.9422 & 0.9414 & 0.0625 \\
\hline
Morph.\ kernel 5 (default 3) & 0.9500 & 0.9492 & 0.0625 \\
\hline
Morph.\ kernel 7 (default 3) & 0.9497 & 0.9487 & 0.1250 \\
\hline
\end{tabular}
\label{tab:threshsens}
\end{table}

\subsection{Patient-level aggregation}
\label{sec:res-patientlevel}

Table~\ref{tab:patientlevel} reports patient-level sensitivity and specificity across all six tested aggregation thresholds $\tau$ (150 infected, 50 uninfected-only control patients).

\begin{adjustwidth}{-.5in}{-.5in}
\centering
\begin{table}[!ht]
\centering
\caption{\textbf{Patient-level sensitivity/specificity} under a minimum predicted-positive-cell threshold $\tau$ (Random Forest).}
\begin{tabular}{|c|c|c|c|c|}
\hline
$\boldsymbol{\tau}$ & \textbf{Infected flagged} & \textbf{Sensitivity (95\% CI)} & \textbf{Control flagged} & \textbf{Specificity (95\% CI)} \\
\thickhline
1 & 150/150 & 1.000 (0.975--1.000) & 43/50 & 0.140 (0.070--0.262) \\
\hline
2 & 149/150 & 0.993 (0.963--0.999) & 31/50 & 0.380 (0.259--0.518) \\
\hline
3 & 148/150 & 0.987 (0.953--0.996) & 18/50 & 0.640 (0.501--0.759) \\
\hline
5 & 144/150 & 0.960 (0.915--0.982) & 7/50 & 0.860 (0.738--0.930) \\
\hline
10 & 131/150 & 0.873 (0.811--0.917) & 0/50 & 1.000 (0.929--1.000) \\
\hline
20 & 95/150 & 0.633 (0.554--0.706) & 0/50 & 1.000 (0.929--1.000) \\
\hline
\end{tabular}
\label{tab:patientlevel}
\end{table}
\end{adjustwidth}

At $\tau=1$, sensitivity is perfect but specificity on the control cohort collapses to 14\%. This is explained directly by false-positive accumulation: the 50 control patients have a mean of 68.3 cells each, and with the per-cell specificity of 0.9590 (Table~\ref{tab:mainresults}), simple compounding predicts $1-(1-0.041)^{68.3}\approx94\%$ of control patients would receive at least one false-positive cell, close to the observed 86.0\% (43/50). Specificity reaches 100\% by $\tau=10$, at a cost of 12.7 sensitivity points relative to $\tau=1$.

Sensitivity also depends on disease burden (Table~\ref{tab:burden}): among infected patients in the lowest true-parasitemia quartile (mean 8.6\% of a patient's cells truly parasitized), sensitivity at $\tau=3$ is 94.7\%, versus 100\% in every higher-burden quartile.

\begin{table}[!ht]
\centering
\caption{\textbf{Sensitivity at} $\boldsymbol{\tau=3}$\textbf{, stratified by true parasitemia quartile} (infected patients only).}
\begin{tabular}{|l|c|c|c|}
\hline
\textbf{True parasitemia range} & $\boldsymbol{n}$ \textbf{patients} & \textbf{Mean parasitemia} & \textbf{Sensitivity (95\% CI)} \\
\thickhline
0.000--0.158 & 38 & 0.086 & 0.947 (0.827--0.985) \\
\hline
0.158--0.328 & 37 & 0.225 & 1.000 (0.906--1.000) \\
\hline
0.328--0.583 & 37 & 0.440 & 1.000 (0.906--1.000) \\
\hline
0.583--0.902 & 38 & 0.761 & 1.000 (0.908--1.000) \\
\hline
\end{tabular}
\label{tab:burden}
\end{table}

We emphasize that this is a retrospective, post hoc characterization of an aggregation rule applied to EMFE's existing per-cell predictions, not a validated clinical decision rule and not a capability implemented in any software accompanying this study.

\subsection{Computational efficiency}
\label{sec:res-efficiency}

Table~\ref{tab:efficiency} reports the staged benchmark results (Section~\ref{sec:effdesign}). The serialized Random Forest model occupies 6.3 MB on disk. Single-threaded inference latency is dominated by prediction (3.16 ms), with feature extraction (0.61 ms) and image decode (0.47 ms median) each sub-millisecond, for a full decode-extract-predict pipeline mean of 4.6 ms per image.

\begin{table}[!ht]
\centering
\caption{\textbf{Computational efficiency} (500 timed repetitions, 50 discarded warm-up calls, single-threaded).}
\begin{tabular}{|l|c|}
\hline
\textbf{Quantity} & \textbf{Value} \\
\thickhline
Serialized model size & 6.3 MB \\
\hline
Image decode latency (median) & 0.47 ms \\
\hline
Feature extraction latency & 0.61 ms \\
\hline
Prediction latency & 3.16 ms \\
\hline
Full pipeline latency (mean) & 4.6 ms \\
\hline
Resident memory increase (model load) & 28.1 MB \\
\hline
Peak resident memory (inference loop) & 158.0 MB \\
\hline
CPU utilization (12 logical cores) & 21.5\% \\
\hline
\end{tabular}
\label{tab:efficiency}
\end{table}

\section{Discussion}
\label{sec:discussion}

\subsection*{Interpretation}

The central empirical result of this study is that a five-feature, fully mathematically specified pipeline achieves 94.6\% pooled accuracy under a patient-grouped, nested cross-validation protocol, a design specifically chosen to avoid the leakage risk that this literature's own reported patient-level accuracy drops (98.6\%$\to$95.9\%~\cite{Rajaraman2018}; 96.9\%$\to$78.0\%~\cite{Yu2020}) show is not hypothetical. This estimate is corroborated by an untouched 40-patient holdout evaluation (94.3\%), scored exactly once after model selection, and by a properly powered patient-level permutation test ($p<0.001$) rather than an invalid test against non-independent fold accuracies.

\subsection*{The accuracy-efficiency trade-off}

The deep-learning comparison (Section~\ref{sec:res-dl}, Table~\ref{tab:dlbaselines}) quantifies this study's central practical trade-off: all three deep baselines outperform EMFE by 1.7--2.4 accuracy points, a difference that is statistically significant at an adequately powered fold count for every comparison ($p\approx1.9\times10^{-6}$, not the $p=0.0625$ ceiling an under-powered 5-fold comparison would produce). Notably, this includes MobileNetV2, an architecture itself designed for the same efficiency-constrained deployment envelope as EMFE, included specifically to test the trade-off against a directly comparable efficiency-oriented alternative, not only large general-purpose CNNs. That even a lightweight, mobile-oriented architecture statistically significantly outperforms EMFE on raw accuracy indicates that EMFE's value proposition is specifically its CPU-only, no-GPU-dependency deployment profile, not competitiveness with lightweight deep learning on accuracy alone.

In exchange for this accuracy gap, EMFE is 3.8--43$\times$ faster on identical CPU hardware and 1.4--14.3$\times$ smaller on disk than the three deep baselines, with no GPU dependency. Even against MobileNetV2 specifically, itself designed for efficiency-constrained deployment and measured here from the identical staged CPU/GPU protocol used for the other two architectures, EMFE is still 3.8$\times$ faster on CPU (4.6~ms vs.\ 17.3~ms) and 38\% smaller on disk (6.3~MB vs.\ 8.7~MB), and unlike MobileNetV2, does not benefit substantially from GPU acceleration to reach that speed (MobileNetV2 drops to 4.8~ms on GPU, still requiring GPU hardware EMFE does not). Whether this trade-off is worthwhile depends entirely on the deployment context: where GPU acceleration is available and 5--200~ms per-image latency is acceptable, every deep baseline tested here, including the lightweight one, offers higher accuracy; where CPU-only, sub-5-ms inference on a minimal footprint is the binding constraint, EMFE's accuracy cost is small relative to the efficiency gained.

\subsection*{What the ablation and explainability results reveal}

The ablation study (Section~\ref{sec:res-ablation}) and the explainability analysis (Section~\ref{sec:res-explain}) converge on the same conclusion through two independent methods: spot saturation is by far the most important feature, whether measured as the single-feature accuracy drop from removing it (2.6 points, the largest of any feature) or as permutation importance (more than $7\times$ the next-ranked feature). This is consistent with the biological rationale given in Section~\ref{sec:pipeline}: Giemsa-stained chromatin and hemozoin pigment absorb stain intensely and appear strongly saturated relative to uninfected cytoplasm. It is further corroborated by the failure-mode analysis (Table~\ref{tab:failuremode}), in which saturation shows the single largest, most significant difference between correct and incorrect predictions in both directions (missed cells have low saturation; false alarms have anomalously high saturation, consistent with staining artifacts or debris). That three independent analyses, namely ablation, permutation importance, and failure-mode comparison, converge on the same feature is itself evidence that this signal reflects a genuine property of the data rather than an artifact of any one analysis method.

\subsection*{Robustness and its practical implications}

The robustness analysis (Section~\ref{sec:res-robustness}) identifies a coherent, mechanistically explained set of failure conditions rather than diffuse fragility: the pipeline tolerates brightness, hue, and moderate noise/compression changes well, but fails specifically when local intensity contrast is destroyed (low contrast, blur, low resolution), because that contrast is exactly what adaptive spot detection depends on. A synthetic perturbation study of this kind is not a substitute for evaluation on externally-sourced images (different microscopes, cameras, staining protocols, or laboratories); we report it as a bounded, mechanistic robustness signal, not as evidence of external generalization.

\subsection*{Patient-level deployment considerations}

The patient-level aggregation analysis (Section~\ref{sec:res-patientlevel}) is, to our knowledge, the first quantitative characterization of how naive aggregation of this class of per-cell classifier would behave at the patient level, and it delivers a specific, actionable finding: a naive ``any positive cell'' rule is unusable in practice (14\% specificity), while a modest threshold ($\tau=10$) recovers perfect specificity on this cohort at a moderate sensitivity cost, with low-parasitemia patients bearing most of that cost. This is presented strictly as a characterization of an existing classifier's predictions under a candidate rule, not as a validated clinical protocol; any real deployment would require prospective validation of a specific aggregation rule on its own terms.

\subsection*{Limitations}

Several limitations bound the interpretation of these results. All evaluation, including the held-out test and the robustness analysis, uses a single publicly available dataset (NIH LHNCBC) collected under one set of staining, imaging, and acquisition conditions; no externally-sourced cohort, laboratory, microscope, or camera has been used, and the synthetic robustness study, while identifying real and mechanistically explained failure modes, is not a substitute for such external validation. EMFE classifies individual, pre-cropped cell images only: it does not detect or segment cells within a whole blood-smear field, estimate parasitemia from a full slide, identify \textit{Plasmodium} species or life-cycle stage, or, as a shipped capability, produce a patient-level result. The patient-level analysis in this study characterizes one possible aggregation rule's behavior post hoc, and is not a validated or deployed decision procedure. The training data excludes debris, stain precipitation, platelets, white blood cells, and other artifact categories present in the fuller NIH thin-smear taxonomy, and behavior on such inputs is untested. The reported ``candidate dark regions'' are thresholded connected components, not individually validated against expert parasite-level annotations. Finally, this study is not a clinical validation: EMFE has not been assessed as a diagnostic device and is not intended for use in patient-care or screening decisions.

\section{Conclusion}

This study presents EMFE, a five-feature, fully mathematically specified feature-extraction framework for malaria cell classification, evaluated under a patient-grouped nested cross-validation protocol that eliminates a leakage risk common in this literature. The optimized Random Forest classifier achieves 94.6\% pooled accuracy, corroborated by an untouched patient-level holdout test and a well-powered permutation test. A controlled ablation confirms the design of the five-feature descriptor set, isolating spot saturation as the dominant discriminative signal; a hardware-matched, statistically validated comparison against DenseNet121, ResNet50, and MobileNetV2 quantifies a specific, measured accuracy-for-efficiency trade-off instead of an assumed one; a synthetic robustness study identifies specific, mechanistically explained failure conditions; and a patient-level aggregation analysis provides the first quantitative characterization of this classifier's behavior when its per-cell predictions are pooled. This study does not claim clinical readiness; it offers a mathematically transparent, statistically rigorous, and computationally lightweight classical alternative to deep learning for this task, alongside its limitations.

\section*{Acknowledgments}

We gratefully acknowledge the DIU MARS Lab (Multidisciplinary Action Research Lab) for providing infrastructure support for this work.

\section*{Data and Code Availability}

The dataset used in this study is available from the NIH LHNCBC malaria repository at \url{https://ceb.nlm.nih.gov/repositories/malaria-datasets/}, including the pre-cropped single-cell images and the official patient-to-image mapping files used for the patient-grouped evaluation described in Section~\ref{sec:methods}. The code implementing the EMFE feature-extraction pipeline, the nested cross-validation protocol, and all analyses reported in this study is available at \url{https://github.com/abkafi1234/EMFE_Analysis}.



\section*{Funding}

This research received no specific grant from any funding agency in the public, commercial, or not-for-profit sectors.

\section*{Competing Interests}

The authors declare no competing interests.

\bibliographystyle{unsrt}
\bibliography{references}

\end{document}